%% file: main.tex
\documentclass{article}
\usepackage{INTERSPEECH2019,amsmath,graphicx,tabularx,blindtext,color,verbatim}

\title{Lattice-Based Unsupervised Test-Time Adaptation\\of Neural Network Acoustic Models}
\name{Ond\v{r}ej Klejch, Joachim Fainberg, Peter Bell, Steve Renals}
\address{
  Centre for Speech Technology Research, University of Edinburgh, Edinburgh EH8 9AB, UK}
\email{\{o.klejch, j.fainberg, peter.bell, s.renals\}@ed.ac.uk}

\begin{document}
\ninept
\maketitle
\begin{abstract}

Acoustic model adaptation to unseen test recordings aims to reduce the mismatch between training and testing conditions.  Most
adaptation schemes for neural network models require the use of an initial one-best transcription for the test data, generated by an unadapted model, in order to estimate the adaptation transform.  It has been found that adaptation methods using discriminative objective functions -- such as cross-entropy loss -- often require careful regularisation to avoid over-fitting to errors in the one-best transcriptions.  In this paper we solve this problem by performing discriminative adaptation using lattices obtained from a first pass decoding, an approach that can be readily integrated into the lattice-free maximum mutual information (LF-MMI) framework.  We investigate this approach on three transcription tasks of varying difficulty: TED talks, multi-genre broadcast (MGB) and a low-resource language (Somali).  We find that our proposed approach enables many more parameters to be adapted without over-fitting being observed, and is successful even when the initial transcription has a WER in excess of 50\%.
\end{abstract}

\section{Introduction}
\label{sec:intro}

Acoustic model adaptation aims to improve automatic speech recognition (ASR) accuracy  by reducing the mismatch between training and test conditions.   In feature-space adaptation,  transformations of acoustic features are estimated to maximise the log-likelihood of the adaptation data~\cite{leggetter1995maximum,gales1998maximum}.  A subset of the weights of a neural network acoustic model \cite{swietojanski2016learning, xue2014singular, liao2013speaker, yu2013kl} are adapted in model-based adaptation. Hybrid adaptation uses auxiliary features such as i-vectors~\cite{dehak2011front,saon2013speaker} or speaker codes~\cite{abdel2013fast} to inform the acoustic model about speaker identities. Experiments have shown that these approaches are complementary and can be usefully combined \cite{samarakoon2016combining}. 

A label sequence is provided for the adaptation data in supervised adaptation, but for unsupervised  adaptation only an unlabelled recording is available.  Conventionally, the best path from a first pass decoding is used to estimate labels for unsupervised adaptation~\cite{woodland2001speaker}.  In this paper we focus on unsupervised test-time adaptation of neural network acoustic models. 

An important challenge for unsupervised model adaptation of neural networks is that we do not want to overfit to errors made in the first pass decoding. In the past this challenge was tackled by filtering adaptation data by confidences produced by an ASR system~\cite{woodland2001speaker,mathias2005discriminative,liu2007investigating,walker2017semi,vesely2017semi} or by using an ASR quality estimation~\cite{falavigna2017dnn}. Alternatively, neural network based acoustic models were prevented from overfitting to those errors by limiting expressivity of the adaptation by drastically reducing the number of adapted parameters, for example by adapting only amplitudes of hidden units~\cite{swietojanski2014learning,samarakoon2016subspace} or by using low rank linear transformations~\cite{zhao2017extended}. Furthermore, strong regularisers were used to prevent the outputs or weights of the acoustic model from diverging too far from the original model~\cite{li2006regularized,yu2013kl}. In this paper we explore an alternative approach in which we use a lattice obtained from the first pass decoding as the supervision for unsupervised model adaptation, since lattices contain all information about the ASR system's confidence and possible phone confusions that can be leveraged during adaptation of the acoustic model. Lattices were previously used as supervision for unsupervised adaptation of Gaussian Mixture Models~(GMM) using maximum likelihood linear regression (MLLR) for unsupervised training of GMMs using maximum likelihood training~\cite{fraga2011lattice}. However, in this paper we are interested in unsupervised adaptation of much larger discriminative models. 

An effective unsupervised adaptation technique for neural network acoustic models requires three components. First, it is necessary to select a suitable subset of model parameters for adaptation in order to allow rapid adaptation using small amounts of adaptation data. Second, the system should filter the possible adaptation data with respect to its suitability for adaptation, as the first pass decoding may produce erroneous transcripts. Third, it needs a reliable adaptation schedule that updates the selected adaptation parameters using data filtered by the second component, while preventing overfitting to the adaptation data.  

In this paper we explore an alternative solution to the data filtering component, in which all the adaptation data is used to adapt the whole neural network acoustic model, but the uncertainty in the decoding is captured through the use of complete lattices for supervision. Our approach is inspired by recent work on semi-supervised learning using the sequence level lattice-free maximum mutual information (LF-MMI) objective~\cite{manohar2018semi}, in which it is shown that using lattices as supervision is beneficial compared to using only best paths in the semi-supervised learning setting. We acknowledge that semi-supervised training and test-time adaptation are essentially equal, but we emphasise that this approach allows us to reliably adapt all weights of neural network models using unsupervised adaptation and a discriminative training criterion, which was problematic in the past. Moreover, test-time speaker adaptation uses much less data than semi-supervised training (from 5~minutes to 1~hour) and it uses the same data for adaptation and testing.

We compare the lattice approach to using the best path, both obtained from the first pass decoding for unsupervised adaptation using the LF-MMI framework.  This is experimentally explored using three transcription tasks -- TED talks~\cite{rousseau2012ted,rousseau2014enhancing}, multi-genre TV broadcasts~\cite{bell2015mgb}, and a low-resource language, Somali -- which have a wide range of baseline word error rates (WERs) (from 10\% to 57\%). We demonstrate improvements compared to using only the best path as supervision. Moreover, we show that by using this approach adapting all the parameters achieves better results than commonly used methods that adapt only a small subset of the weights, such as LHUC~\cite{swietojanski2014learning}.

\section{Methods}

\subsection{Lattice supervision and LF-MMI}
Discriminative training using criteria such as maximum mutual information (MMI) \cite{bahl1986maximum} has been shown to be sensitive to the accuracy of the transcripts \cite{mathias2005discriminative,yu2010unsupervised}. In lieu of better transcripts, a range of transcript filtering approaches have previously been explored \cite{mathias2005discriminative,liu2007investigating,walker2017semi}. In unsupervised or semi-supervised approaches, in which we generate hypothesis transcriptions by decoding with a seed model, we can alternatively use a lattice of supervision. For instance, with the MMI criterion:

\begin{equation}
    \mathcal{F}_{MMI}(\lambda) = \sum_{r=1}^R \log \frac{p_\lambda (\mathcal{O}_r | \mathcal{M}_r^{num})}{p_\lambda (\mathcal{O} | \mathcal{M}_r^{den})},
\end{equation}
the numerator lattice $\mathcal{M}^{num}$ can contain multiple hypotheses for the same audio segment $r$ up to some lattice pruning factor. If set to 0, $\mathcal{M}^{num}$ is left with only the best path.

Lattice supervision has previously been used in work on unsupervised adaptation~\cite{padmanabhan2000lattice} and training~\cite{fraga2011lattice} of GMMs, as well as discriminative \cite{povey2005discriminative} and semi-supervised training \cite{manohar2018semi} of neural network models. Following Manohar et al. \cite{manohar2018semi}, we explore the use of lattice supervision versus that of only using the best path in the denominator lattice-free version of MMI (LF-MMI) \cite{povey2016purely}. LF-MMI  was introduced by Povey et al. \cite{povey2016purely} as a method to train neural network acoustic models with a sequence discriminative criterion (MMI) without an initial cross-entropy (CE) stage to generate lattices approximating all possible word sequences (e.g. \cite{vesely2013sequence}). The word-level denominator lattice is instead replaced with a phone-level denominator graph encoding all possible sequences given a 3 or 4-gram phone language model. To further reduce complexity, the model outputs at one third of the frame rate. In the numerator a frame-by-frame mask allows phones to appear with some tolerance relative to its original alignment. A mixture of regularisation methods are required to reduce overfitting; for more details we refer to \cite{povey2016purely}. Povey et al. \cite{povey2016purely} demonstrated up to 8\% relative improvements in WER over previous CE systems followed by sequence discriminative training with the state Minimum Bayes Risk (sMBR) criterion~\cite{kaiser2000novel}. An extension to the LF-MMI framework was recently proposed that enables flat-start training with neural networks \cite{hadian2018end}.


\subsection{LHUC}

Traditionally, only a subset of the acoustic model weights is adapted to prevent overfitting to the transcripts obtained from a first pass decode. This is usually done by inserting a linear layer after the input layer, hidden layers or output layers. For example, an adaption of activations $h$ of a hidden layer with speaker dependent weights $A$, can be expressed as follows:

\begin{equation}
    h' = A \cdot h.
\end{equation}

However, even these simple techniques tend to overfit because too many parameters are used for adaptation.  Learning Hidden Unit Contributions (LHUC) \cite{swietojanski2014learning,swietojanski2016learning} is a technique in which only elements on the diagonal of the speaker dependent matrix $A$ are adapted -- i.e.\ each hidden unit maybe viewed as having a speaker adaptive amplitude parameter. Since a much smaller number of weights is adapted, this technique is not that prone to overfitting to adaptation data.  

LHUC was developed in the context of frame-based hybrid neural network / hidden Markov models, trained using the cross-entropy criterion.  It has previously been successfully applied to sequence discriminatively trained models by using cross-entropy for the LHUC updates \cite{swietojanski2016learning}.

\subsection{SAT-LHUC}

A way to further improve model adaptation is to use speaker adaptive training \cite{anastasakos1996compact}. Therefore, we also trained a SAT-LHUC model~\cite{swietojanski2016sat,swietojanski2016learning}. This was done by maintaining speaker-dependent LHUC parameters for each speaker and training them jointly with speaker-independent parameters. In order to obtain a good speaker-independent model that can be used for decoding and adaptation to new speakers, we trained speaker-independent LHUC parameters instead of speaker dependent with probability 0.5 during training. This is similar to a  setup that was shown to well in~\cite{swietojanski2016learning}. At the beginning of training all speaker dependent parameters were sampled from a normal distribution with $\mu=1$ and $\sigma=0.01$. It was important to turn off L2 regularisation and parameter shrinkage for the speaker dependent parameters in Kaldi, as otherwise all speaker dependent parameters would converge to zero. 

\section{Experiments}

We conducted test-time model adaptation experiments on three datasets: the TED-LIUM corpus of TED talks \cite{rousseau2012ted,rousseau2014enhancing}, multi-genre TV broadcasts from the MGB~1 Challenge~\cite{bell2015mgb} and a corpus of Somali from the IARPA MATERIAL programme. All models were trained and adapted using the Kaldi toolkit~\cite{povey2011kaldi}. We describe the respective baseline models in sections 3.1-3.3 and the adaptation of the models in 3.4.

\subsection{TED-LIUM}

We trained three time-delayed neural network (TDNN) models~\cite{peddinti2015time} with LF-MMI~\cite{povey2016purely} following Kaldi TED-LIUM recipe 1f. All models had the same architecture with 7 hidden layers with 450 units. The first model was trained without i-vector features, the second model was trained with i-vector features and the third model was trained without i-vector features but with SAT-LHUC~\cite{swietojanski2016sat}. All models were trained only on TED talks that were recorded before 2012 in order to conform with the IWSLT~\cite{iwslt2012} evaluation guidelines which resulted in 130 hours of training data. We performed adaptation on the TED-LIUM dev and test data which contained 8 speakers with an average speech duration of 11.9 minutes and 11 speakers with an average speech duration of 14.2 minutes respectively.

\subsection{MGB}\label{sec:mgb}
For the MGB-1 corpus \cite{bell2015mgb} we trained a factored TDNN (TDNN-F) model~\cite{povey2018semi} with LF-MMI following Kaldi Switchboard recipe 7p. The model has 12 layers with 1280 units each (apart from the penultimate layer) and a bottleneck dimension of 256, with batch normalisation and dropout layers interleaved throughout. The model was trained for 8 epochs. We used alignments obtained with the HMM-GMM recipe provided with the MGB challenge, and trained on transcripts from a lightly supervised decode that had a maximum matching error rate (MER) with the original subtitles of 40. This yields roughly 649 hours of data, or 1960 hours after speed perturbation. For adaptation we carry across all training parameters (including dropout which we found particularly important for this data) and we rescore the supervision lattices with a 4-gram language model (LM) as in \cite{manohar2018semi}, estimated on about 640 million words of BBC subtitle text. We adapt and test on the longitudinal eval set which consists of 10 hours across two TV shows and a total of 19 episodes, each between 30 and 45 minutes in length. We do not have speaker clustering and therefore extract i-vectors per utterance and perform episode level adaptation. Finally, we rescore the decoded output with the 4-gram LM.

\subsection{Somali}

We carried out experiments on Somali ``surprise language'' data released to participants on the IARPA-MATERIAL programme\footnote{https://www.iarpa.gov/index.php/research-programs/material}.  Training data comprises 499 narrow-band telephone conversations sides, totalling 37 hours of speech.  
Test data comprises narrowband telephone conversations (NB); and wideband (WB) data from the news and topical broadcast domains that are mismatched to the training material.
We trained a TDNN-F model using the neural network architecture from Kaldi TED-LIUM recipe 1g. The model had 14 hidden layers with 1024 units. The weight matrices were factored into two matrices with a bottleneck dimension 128. The model used filterbank, pitch and probability of voicing~\cite{ghahremani2014pitch} features together with multilingual bottleneck features obtained from a neural network that was trained on all Babel languages~\cite{cui2016multilingual,gales2017low}. We used per utterance cepstral mean and variance normalisation, since there were no speaker clusters for the wideband test data.

The model was trained on narrowband data with speed perturbation and evaluated on both narrowband and wideband data. We used data scraped from the web to build a language model for wideband data. We performed speaker adaptation on narrowband data which consisted of 117 speakers with an average speech duration of 4.7 minutes and file-level adaptation on wideband data which consisted of 119 files with an average speech duration of 5 minutes.


\subsection{Adaptation methods}

In this paper we were primarily interested in comparing model adaptation methods that use either one best path (called \textbf{BP} in the Results section) or a lattice (called \textbf{LAT} in the Results section) obtained from the first pass decoding for supervision. We adjusted a recipe for semi-supervised training using LF-MMI~\cite{manohar2018semi} to instead perform test-time adaptation. Our main hypothesis was that methods using lattices for supervision are much less likely to overfit to incorrectly transcribed segments in the adaptation data. In the past when only the best path was used for model adaptation, several techniques for data selection were required~\cite{mathias2005discriminative,liu2007investigating,walker2017semi}. In this paper we compared adapting using only utterances with top $25\%$, $50\%$ or $75\%$ average utterance confidence. We conducted model adaptation experiments in two regimes: in the first regime, we adapted all parameters of the acoustic model (called \textbf{ALL} in the Results section); in the second regime, we adapted only LHUC parameters inserted after every hidden layer of the acoustic model (called \textbf{LHUC} in the Results section). When adapting all parameters, we adapted the model for three epochs, starting with the learning rate which was used in the last iteration during training. We gradually decreased the learning rate down to one tenth (one fifth for MGB) of the initial learning rate. This learning schedule was chosen in order to imitate continued learning of the model. When adapting LHUC parameters, we adapted the model for three epochs with a fixed learning rate of $0.7$, which we found to work well in previous experiments.

\section{Results}

\input{tables/tedlium_1.tex}
\input{tables/tedlium_2.tex}
\input{tables/tedlium_3.tex}


We conducted the first set of experiments on the TED-LIUM dataset. Adaptation of the model without i-vectors using lattices achieves $10-15\%$ relative improvement when adapting LHUC parameters and $9-14\%$ relative improvement when adapting all parameters, whereas improvements when adapting using best path and all adaptation data were much smaller or even negative (Table~\ref{tab:ted_no_ivectors}). We observed a similar trend for the remaining two models. Adaptation of the model using i-vectors (Table~\ref{tab:ted_ivectors}) using lattices as supervision improves performance of a speaker adaptive baseline. This confirms that model-based adaptation is complementary with i-vectors. Unfortunately, our implementation of SAT-LHUC (Table~\ref{tab:ted_sat_lhuc}) did not outperform test-only LHUC adaptation. We plan to explore other possibilities of SAT-LHUC training in the future.

\input{tables/mgb.tex}
For the MGB corpus we adapted to entire episodes in the longitudinal eval set, rather than to speakers, as noted in Section~\ref{sec:mgb}. This provides more adaptation data (30-45 minutes per episode), but perhaps at the cost of losing finer granularity for adaptation. Adapting all parameters with lattice supervision provided the best results (Table~\ref{tab:mgb}). This is to our knowledge the best results shown on the longitudinal evaluation set \cite{bell2015mgb}. Using the best path with all parameters yields almost no gains ($\sim1\%$). When only adapting a subset of the parameters with LHUC the results are more stable, but does not perform as well as all parameters with lattice supervision.

\input{tables/somali.tex}

We also evaluated adaptation using lattices as supervision on the Somali data. As can be seen from the table, Somali data is very challenging -- the initial WER of the model is very high on both NB and WB data at $53.7\%$ and $57.3\%$ respectively. 
These results are similar to other experiments conducted on other IARPA-MATERIAL programme languages with the same TDNN-F neural network architecture~\cite{povey2018semi}. Here we show that adapting such a model using a best path as supervision does not reduce the WER, because the best path contains too many errors. Nevertheless, adaptation using lattices as supervision gives $0.7-0.8\%$ absolute improvements. Even though the relative improvement is small, it is interesting to see that using lattices as supervision allows us to improve performance at all. We believe that adapting to entire files is sub-optimal, because the speaker variance in the wide-band data might be too high. Therefore, we plan to perform per utterance adaptation experiments in the future.

\input{tables/filtered_data.tex}

One common way to prevent adaptation to erroneous first pass transcripts is to filter the adaptation data by confidences~\cite{vesely2017semi}, for example by the average utterance confidence.
This filtering can be done by using a hard threshold, or by using only the fraction of utterances with the highest confidences. Either way one extra hyper-parameter is introduced. In Table~\ref{tab:filtered_data} we compare adaptation using lattices as supervision with adaptation using only best paths on various fractions of the adaptation data when adapting all parameters. We experiment with the TED-LIUM model without i-vectors, and the Somali model. As can be seen from the table, filtering utterances improves results when using best path supervision. The biggest improvement can be achieved when using only $50\%$ of the adaptation data. Even then the TED-LIUM model does not obtain similar performance as when adapted using lattices for supervision. Furthermore, adaptation of the Somali model using best path supervision only barely matches the unadapted baseline. This is probably due to the fact that the WER of the initial Somali model is high and that the lattice provides much more information than a combination of best path supervision and corresponding confidences. We also performed the same filtering experiment with lattices as supervision. We found that using a threshold of 75\% -- 100\% achieves the best results. Overall, adaptation using lattice supervision does not benefit from filtering utterances as much as adaptation using best path supervision. 


\section{Conclusions}

In this paper we compared unsupervised model adaptation using a lattice with the best path obtained from the first pass decoding as supervision.  Our experiments show that using the lattice as supervision achieves better results than using the best path, even when confidence-based data selection is used to remove transcripts with many possible errors. This is due to the fact that the lattice from the first pass decoding contains much more information, such as confidence and phonetic confusions, than the best path. We find that the use of a lattice as supervision is particularly important when adapting all parameters, when over-fitting to incorrect first-pass transcriptions is a particular problem: in many cases we outperform a strong baseline that adapts only LHUC parameters. Moreover, we showed that when using lattices as supervision it is possible to adapt a model whose initial WER is higher than $50\%$, for which adapting with best path supervision often produced worse WERs than the unadapted baseline.

Our finding that use of lattices greatly aids the two adaptation methods we considered motivates further investigation into whether other test-time adaptation
techniques -- many of which show limited gains in an unsupervised setting -- could benefit similarly.  This will be the subject of further work.

\medskip\noindent
\textbf{Acknowledgements:}
This research is based upon work supported in part by the Office of the Director of National Intelligence (ODNI), Intelligence Advanced Research Projects Activity (IARPA), via Air Force Research Laboratory (AFRL) contract \#FA8650-17-C-9117. The views and conclusions contained herein are those of the authors and should not be interpreted as necessarily representing the official policies, either expressed or implied, of ODNI, IARPA, AFRL or the U.S. Government. The U.S. Government is authorized to reproduce and distribute reprints for governmental purposes notwithstanding any copyright annotation therein. This work was also partially supported by: the EU H2020 projects SUMMA (grant agreement 688139) and ELG (grant agreement 825627), and a PhD studentship funded by Bloomberg.

\bibliographystyle{IEEEtran}
\bibliography{main}

\end{document}

%% file: tables/tedlium_1.tex
\begin{table}[t!]
\begin{center}
\caption{WER for adaptation of the TED-LIUM model without i-vectors.}
\label{tab:ted_no_ivectors}
\begin{tabularx}{\columnwidth}{Xcc}
\toprule
					        	& 	dev		&	test	\\
\midrule
\textbf{baseline}				&	10.0 	&	10.6 	\\
\textbf{LHUC-LAT}		    	&	 9.0	&	9.3 	\\
\textbf{LHUC-BP}				&	 9.8	&	10.1	\\
\textbf{ALL-LAT}				& 	 9.1	&	 9.0	\\
\textbf{ALL-BP}  				& 	 9.9	&	10.6	\\
\bottomrule
\end{tabularx}
\end{center}
\end{table}

%% file: tables/tedlium_2.tex
\begin{table}[t!]
\begin{center}

\caption{WER for adaptation of the TED-LIUM model using i-vectors.}
\label{tab:ted_ivectors}

\begin{tabularx}{\columnwidth}{Xcc}
\toprule
					        	& 	dev		&	test	\\
\midrule
\textbf{baseline}				&	 9.0 	&	 9.5 	\\
\textbf{LHUC-LAT}		    	&	 8.8	&	 8.9 	\\
\textbf{LHUC-BP}				&	 9.1	&	 9.3	\\
\textbf{ALL-LAT}				& 	 8.8	&	 8.9	\\
\textbf{ALL-BP}  				& 	 9.1	&	 9.6	\\
\bottomrule
\end{tabularx}
\end{center}
\end{table}

%% file: tables/tedlium_3.tex
\begin{table}[t!]
\begin{center}
\caption{WER for adaptation of the TED-LIUM SAT-LHUC model.}
\label{tab:ted_sat_lhuc}

\begin{tabularx}{\columnwidth}{Xcc}
\toprule
					        	& 	dev		&	test	\\
\midrule
\textbf{baseline}				&	 9.7 	&	10.2 	\\
\textbf{LHUC-LAT}			    &	 9.4	&	 9.1 	\\
\textbf{LHUC-BP}				&	 9.7	&	 9.8	\\
\textbf{ALL-LAT}				& 	 9.2	&	 8.9	\\
\textbf{ALL-BP}  				& 	 9.8	&	10.4	\\
\bottomrule
\end{tabularx}
\end{center}
\end{table}

%% file: tables/mgb.tex
\begin{table}[t!]
\begin{center}
\caption{WER for adaptation of the MGB model to episodes in the longitudinal eval data.}
\begin{tabularx}{\columnwidth}{Xc}
\toprule
					        	&	eval	\\
\midrule
\textbf{baseline}				&	19.9 	\\
\textbf{LHUC-LAT}			&	19.4  	\\
\textbf{LHUC-BP}				&	19.5 	\\
\textbf{ALL-LAT}				&	19.2 	\\
\textbf{ALL-BP}  				&	19.7 \\
\bottomrule
\end{tabularx}
\label{tab:mgb}
\end{center}
\end{table}


%% file: tables/somali.tex
\begin{table}[t!]
\begin{center}
\caption{WER for adaptation of the Somali model on narrow-band (NB) dev data and wide-band (WB) test data.}
\begin{tabularx}{\columnwidth}{Xcc}
\toprule
					        	& 	NB		&	WB	\\
\midrule
\textbf{baseline}				&	53.7 	&	57.3 	\\
\textbf{LHUC-LAT}		    	&	53.6	&	56.7 	\\
\textbf{LHUC-BP}				&	54.1	&	57.9	\\
\textbf{ALL-LAT}				& 	53.0	&	56.5	\\
\textbf{ALL-BP}  				& 	54.5	&	58.2	\\
\bottomrule
\end{tabularx}

\label{tab:somali}
\end{center}
\end{table}

%% file: tables/filtered_data.tex
\begin{table}[t!]
\begin{center}
\caption{WER for adaptation of the TED-LIUM model without i-vectors and the Somali model using best path as a supervision with varying fractions of the adaptation data.}
\begin{tabularx}{\columnwidth}{Xcccccc}
\toprule
                                & \multicolumn{2}{c}{TED-LIUM} & \multicolumn{2}{c}{Somali} \\
					        	& dev   & test  & NB   & WB	\\
\midrule
\textbf{baseline}				& 10.0  & 10.6  & 53.7 	& 57.3 	\\
\textbf{ALL-LAT 100\%}          &  9.1  &  9.0  & 53.0  & 56.5  \\
\textbf{ALL-LAT 75\%}           &  9.2  &  8.8  & 53.3  & 56.2  \\
\textbf{ALL-LAT 50\%}           &  9.4  &  9.0  & 53.8  & 56.5  \\
\textbf{ALL-LAT 25\%}           &  9.7  &  9.5  & 56.0  & 57.0  \\
\textbf{ALL-BP 100\%}           &  9.9  & 10.6  & 54.5  & 58.2  \\
\textbf{ALL-BP 75\%}            &  9.6  &  9.7  & 53.8  & 57.8  \\
\textbf{ALL-BP 50\%}            &  9.4  &  9.4  & 53.7  & 57.1  \\
\textbf{ALL-BP 25\%}            &  9.6  &  9.6  & 56.0  & 57.2  \\
\bottomrule
\end{tabularx}
\label{tab:filtered_data}
\end{center}
\end{table}